\documentclass[9pt,twocolumn,twoside]{osajnl}

\usepackage{subfig}
\usepackage{float, subfloat}

\DeclareMathAlphabet{\mathsfsl}{OT1}{cmss}{m}{sl}
\newcommand{\matr}[1]{\ensuremath{\mathsfsl{#1}}}

\journal{ol} 

\setboolean{shortarticle}{true} 

\title{Numerical Demultiplexing of Color Image Sensor Measurements via Non-linear Random Forest Modeling}

\author[1*]{Jason Deglint}
\author[1]{Farnoud Kazemzadeh}
\author[1]{Daniel Cho}
\author[1]{David A. Clausi}
\author[1]{Alexander Wong}

\affil[1]{Vision and Image Processing Research Group, University of Waterloo, Waterloo, Ontario, Canada}

\affil[*]{Corresponding author: jdeglint@uwaterloo.ca}
\dates{Compiled \today}

\ociscodes{(100.2000) Digital image processing; (100.3190) Inverse problems; (110.4234) Multispectral and hyperspectral imaging; (150.1708) Color inspection; (300.6550) Spectroscopy, visible.}
\doi{}

\begin{abstract}
The simultaneous capture of imaging data at multiple wavelengths across the electromagnetic spectrum is highly challenging, requiring complex and costly multispectral image sensors.  In this study, we introduce a comprehensive framework for performing simultaneous multispectral imaging using conventional image sensors with color filter arrays via numerical demultiplexing of the color image sensor measurements. A numerical forward model characterizing the formation of sensor measurements from light spectra hitting the sensor is constructed based on a comprehensive spectral characterization of the sensor.  A numerical demultiplexer is then learned via non-linear random forest modeling based on the forward model.  Given the learned numerical demultiplexer, one can then demultiplex simultaneously-acquired measurements made by the image sensor into reflectance intensities at discrete selectable wavelengths, resulting in a higher resolution reflectance spectrum.  Simulation and real-world experimental results demonstrate the efficacy of such a method for simultaneous multispectral imaging.
\end{abstract}

\setboolean{displaycopyright}{true}

\begin{document}

\maketitle
\thispagestyle{fancy}
\ifthenelse{\boolean{shortarticle}}{\abscontent}{}

\section{Introduction}
\label{sec:intro}

Multispectral imaging involves the capturing of imaging data of a particular scene or object at multiple wavelengths across the electromagnetic spectrum.
Because different materials and biologicals reflect, transmit, and/or fluoresce at different wavelengths, multispectral imaging becomes a powerful tool to extract additional information about the scene or object that facilitates unique material characterization and classification beyond what can be captured using conventional camera systems.  As such, multispectral imaging has become a widely-used, powerful tool for different applications such as remote sensing~\cite{berni2009thermal, laliberte2011multispectral, calderon2013high}, material analysis~\cite{kim2001hyperspectral, lu2006hyperspectral, FarnoudNewOne}, and microscopy~\cite{hiraoka2002multispectral, brydegaard2009broad, deglint2015virtual, kazemzadeh2015multispectral}.

Traditionally, multispectral imaging has often been performed in a sequential manner, where imaging data is captured at a specific wavelength in the electromagnetic spectrum one after the other. Such sequential multispectral imaging systems typically consists of a monochromatic sensor and a spectral filtering mechanism such as filter wheels~\cite{yamaguchi2005multispectral} and tunable filters~\cite{gupta2005acousto, harris1969acousto} that allow the desired wavelength of light to pass through for acquisition.  While highly useful for imaging static phenomena in a controlled environment, there are a number of limitations associated with such sequential multispectral imaging systems.  First, such imaging systems require a complex optical setup involving many optoelectronic elements, leading to a more expensive and less compact system.  Second, because imaging data at different wavelengths are captured in a sequential manner, the temporal resolution of such systems is reduced and as such reduce their effectiveness for imaging fast, transient phenomena.

To address such limitations, there has been an on-going trend towards simultaneous multispectral imaging systems, where the goal is to capture imaging data at all desired wavelengths at the same time.  Such systems allow for effective imaging of fast, transient phenomena, and facilitates a less complex, and more compact design.  A major limitation of simultaneous multispectral imaging systems revolve around the use of customized multispectral image sensors, where different sets of pixel sensors in the sensor array are configured to capture imaging data at a particular wavelength.  For example, a multispectral image sensor has been proposed that is capable of capturing eight different wavelengths at the same time~\cite{park2013multispectral}.  However, such custom multispectral image sensors are complex to manufacture and cost-prohibitive for many real-world applications.  As such, a method for simultaneous multispectral imaging that leverages off-the-shelf, low-cost image sensors with standard color filter arrays (CFA) is preferred.

There has been recent interest in exploring simultaneous multispectral imaging systems where higher spectral resolution is obtained from off-the-shelf image sensors with CFAs~\cite{stigell2007Wiener,nishidate2013estimation,chen2012modified,deglint2015inference}.  Autocorrelation and cross-correlation have been used to statistically model the relationship between the incoming light spectra and the image sensor measurements~\cite{stigell2007Wiener,nishidate2013estimation,chen2012modified}, and the resulting autocorrelation and cross-correlation models are used to infer higher resolution reflectance spectra from sensor measurements via Wiener estimation.  One limitation of such an approach is that using only autocorrelation and cross-correlation models limits modeling capabilities for characterizing the complex spectra relationships given inherent statistical assumptions made, and as such may not scale well for obtaining higher spectral resolution.  In this study, inspired by our preliminary work on spectral inference~\cite{deglint2015inference}, we introduce a comprehensive framework for performing simultaneous multispectral imaging using conventional sensors with CFAs via numerical demultiplexing of sensor measurements.  By decoupling these measurements using a low-cost, compact imaging system, one can predict higher resolution spectral data across multiple wavelengths about a scene.

\begin{figure}[]
	\centering
	\includegraphics[width=0.7\columnwidth]{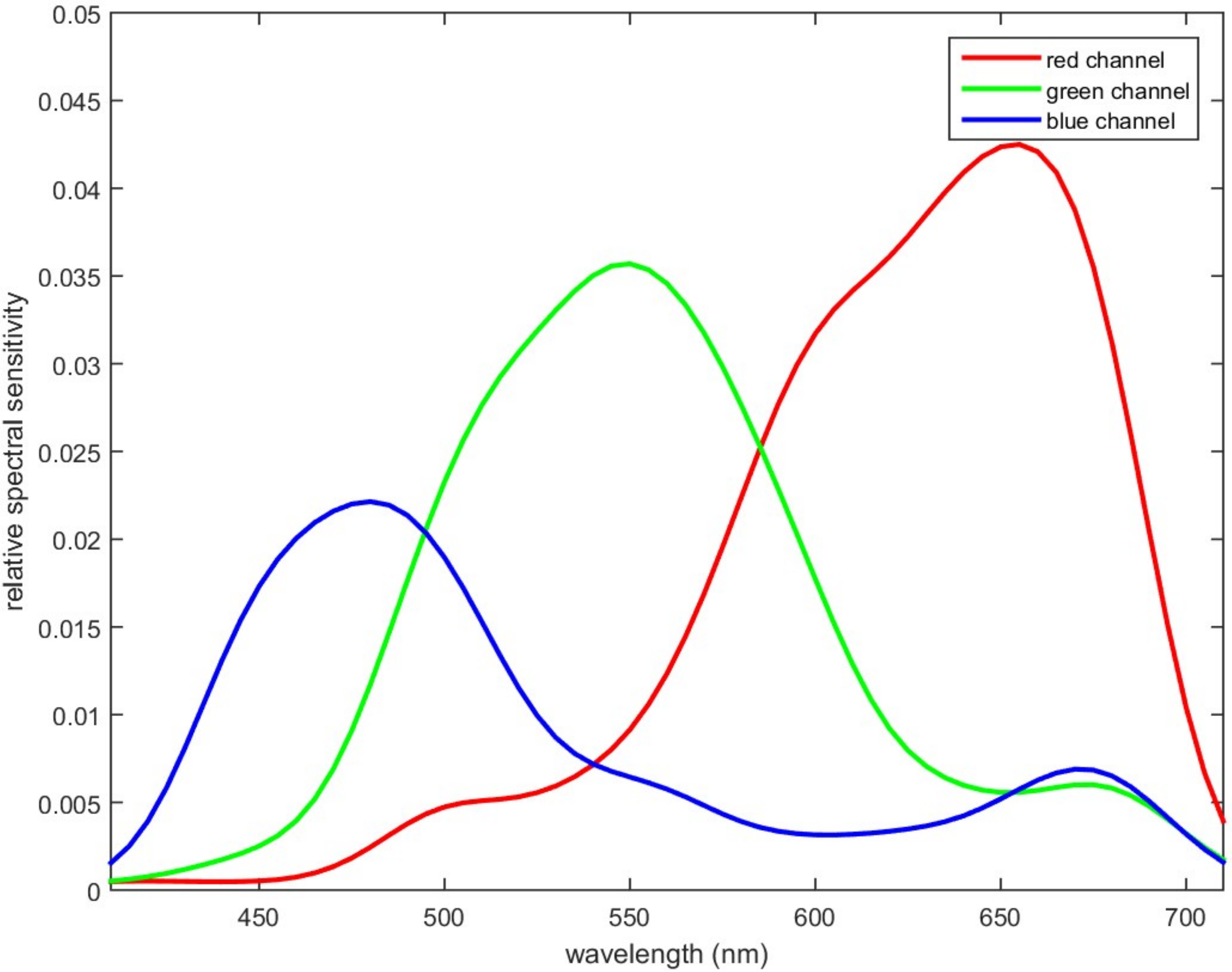} \vspace{-3mm}
	\caption{Spectral response of a Canon T3i APS-C CMOS image sensor with a Bayer pattern CFA.  This spectral characterization is used to construct a forward model characterizing the formation of sensor measurements from light spectra hitting the sensor.}
	\label{fig:spectral_response} \vspace{-3mm}
\end{figure}

\section{Methodology}
\label{sec:method}


The proposed numerical demultiplexing method can be summarized as follows.  First, a comprehensive spectral characterization of the image sensor is performed.  Second, given the spectral characterization information, a forward model is created that maps the input light spectra to the sensor measurements.  Third, given the numerical forward model, the corresponding numerical demultiplexer is constructed via non-linear random forest modeling.  This constructed numerical demultiplexer can then be used to demultiplex simultaneously-acquired measurements made by the image sensor into reflectance intensities at discrete selectable wavelengths, resulting in a higher resolution reflectance spectrum.  A more detailed description of each component of the proposed method is provided below.

\begin{figure}[t!]
	\centering
	\subfloat{\includegraphics[width=0.8\linewidth]{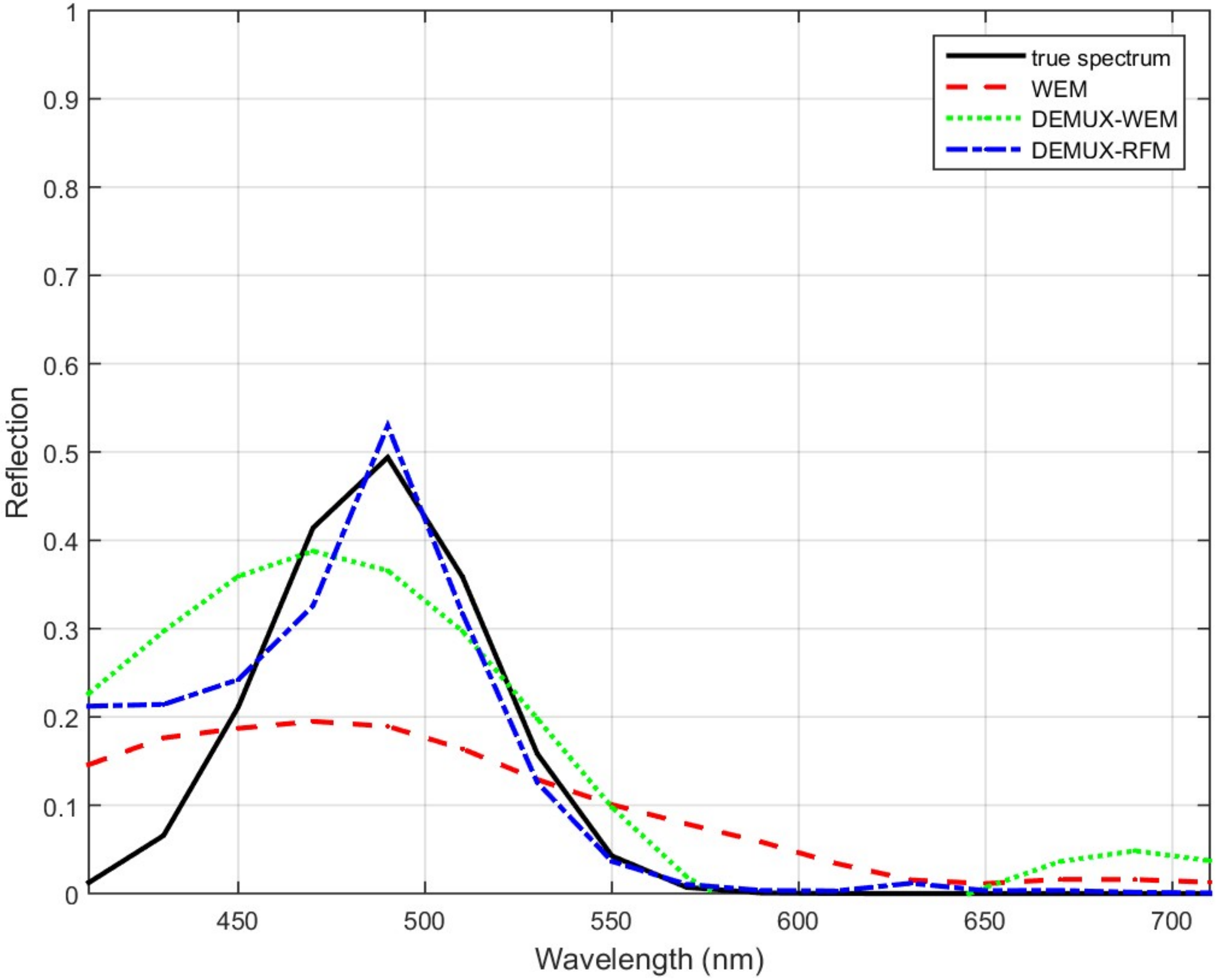}\label{test01_02}}\vspace{-3mm}

	\subfloat{\includegraphics[width=0.8\linewidth]{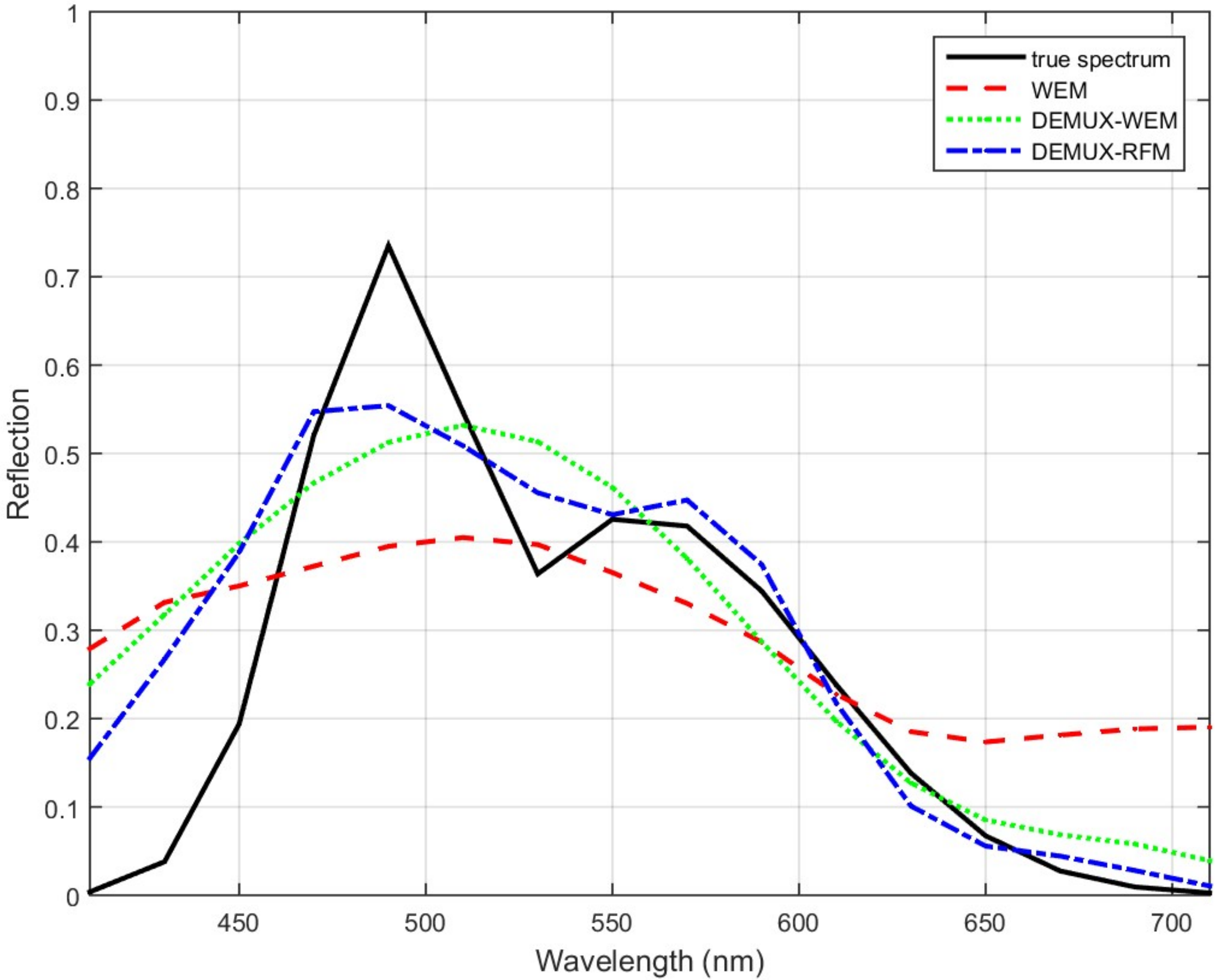}\label{test01_06}}	 \vspace{-3mm}
	
	\caption{Two examples of simulated light spectra that were captured using the simulated sensor to obtain image sensor measurements.  These sensor measurements were then used by the three tested approaches to predict higher resolution reflectance spectra.  The proposed random forest-based demultiplexer (DEMUX-RFM) outperformed both the current state-of-the-art Wiener estimation method (WEM) and the proposed Wiener-based demultiplexer (DEMUX-WEM).}
	\label{fig:virtualspectra}
\vspace{-0.25in}
\end{figure}

\subsection{Spectral Characterization}
In order to construct a numerical forward model characterizing the formation of image sensor measurements given input light spectra, we must first quantitatively characterize the inherent spectral sensitivity of the image sensor.

The intensity measurement on a sensor's pixel, $C(i,j)$,
\begin{equation}
	C(i,j) = \sum_{\lambda} S(i,j,\lambda)R(i,j,\lambda)E(i,j,\lambda),
	\label{eq:spectralSensitivityFunction}
\end{equation}
\noindent depends on the light source $E(i,j,\lambda)$ which is used to illuminate the target, the reflectance of the object $R(i,j,\lambda)$ and the spectral sensitivity of the sensor, $S(i,j,\lambda)$~\citep{sensitivityEquation}, where $(i,j)$ is the pixel location on the sensor and $\lambda$ is the wavelength hitting a given pixel.  The spectral sensitivity, $S(i,j,\lambda)$, can be changed by placing a CFA over the sensor, which facilitates the simultaneous acquisition of multiple spectral bands.

We characterize the spectral sensitivity of a sensor for a given filter in the CFA by emitting a large set of discrete narrowband light spanning the desired wavelength range onto the sensor, and then record the corresponding spectral response within the range of wavelengths.  For example, a very common CFA used in consumer-level color imaging systems is the Bayer pattern CFA~\cite{Bayer76}, which consists of red, green, and blue (RGB) filters placed on the sensor pixels resulting in three-channel spectral measurements. In this study, we designed and built a monochromator that enables wavelength selection of 5 nm and we use the light emerging from the exit slit of the monochromator as an input into the camera to be imaged.
The focus of the camera is placed at infinity to ensure that the beam which impinges on the sensor is as close to collimation as possible, therefore uniformly illuminating a large region on the sensor which results in a large number of each of the three color filters. Using this setup we characterize the spectral sensitivity of a Canon T3i APS-C CMOS image sensor with a Bayer pattern CFA, as described above, using a set of 61 discrete test spectra ranging from 410 nm to 710 nm.   The spectral response curve for the three filters are shown in Fig.~\ref{fig:spectral_response}.


\subsection{Forward Modeling}
Given the spectral characterization of the color sensor, we can now construct a forward model characterizing the color measurement formation by the sensor with a CFA, given input light spectra as follows.  By letting $\Lambda(i,j,\lambda)=R(i,j,\lambda)E(i,j,\lambda)$ Equation~\ref{eq:spectralSensitivityFunction} can be rewritten as
\begin{equation}
	C(i,j) = \sum_{\lambda} S(i,j)\Lambda(i,j,\lambda),
	\label{eq:spectralSensitivityFunction_new}
\end{equation}
\noindent which can be written in matrix form as
$\matr{C}_{p \times 1} = \matr{S}_{p \times n} \matr{\Lambda}_{n \times 1}$.
Here $ \matr{C} = [ c_1  c_2 \hdots c_p ]^T $ represent the measurements made by the image sensor using the $p$ filters in the CFA, $ \matr{\Lambda} = [ \lambda_1  \lambda_2 \hdots \lambda_n ]^T $ represents the intensities at $n$ discrete selectable wavelengths of the light spectra arriving at the sensor, and \matr{S} is the spectral sensitivity of the sensor.  This relationship represents a forward model which maps the light spectra hitting the sensor to the sensor measurements made by the image sensor with a CFA.


\subsection{Numerical Demultiplexer Construction}
At this stage, the goal is to construct a numerical demultiplexer based on the numerical forward model for the characterized image sensor described in Equation~\ref{eq:spectralSensitivityFunction_new}.  One can treat the numerical demultiplexer as an inverse problem of the numerical forward model, with the goal of determining higher resolution reflectance spectra $\matr{\Lambda}$ given the image sensor measurements $\matr{C}$:

\begin{equation}
	\matr{\Lambda} = \matr{S}^{-1}\left(\matr{C}\right)
	\label{eq::inv}
\end{equation}

\noindent where $\matr{S}^{-1}(.)$ is an inverse function that outputs the higher resolution reflectance spectra $\matr{\Lambda}$ given the sensor measurements $\matr{C}$.  Given the complex relationship between the higher resolution reflectance spectra and the sensor measurements, and the fact that we have an underdetermined system in this case, one cannot obtain the inverse function $\matr{S}^{-1}(.)$ in an analytical manner.  Therefore, in the proposed framework, we propose that a numerical demultiplexer can be constructed through nonlinear modeling of the relationship between reflectance spectra and sensor measurements.

More specifically, we leverage non-linear random forest modeling~\cite{Breiman2001} to learn the numerical demultiplexer function $\matr{S}^{-1}(.)$ using a comprehensive dataset consisting of 10000 pairs of reflectance spectra and their corresponding sensor measurements based on the numerical forward model for the characterized image sensor.
A random distribution of reflectance spectra was used in the dataset to ensure that all wavelengths are well represented ensuring that the numerical demultiplexer achieves strong demultiplexing performance across the entire range of wavelengths.  The nonlinear random forest model used in this study for constructing the numerical demultiplexer is comprised of 8000 decision trees in total.   A key advantage of using such a non-linear random forest modeling approach to constructing the numerical demultiplexer is that it allows for reliable and flexible modeling of the complex relationships between reflectance spectra and sensor measurements without imposing strong assumptions about the nature of the relationship.  Furthermore, in this study, we also introduce a numerical Wiener-based demultiplexer based on correlation and autocorrelation models, learned using the numerical forward model for the characterized image sensor described in Equation~\ref{eq:spectralSensitivityFunction_new} for comparison purposes with the proposed random forest-based demultiplexer.

Given the constructed numerical demultiplexer, one can then demultiplex simultaneously-acquired sensor measurements made by the characterized image sensor with a CFA into higher resolution reflectance spectra.

\begin{figure}[t!]
	\centering
	\includegraphics[width=0.75\columnwidth]{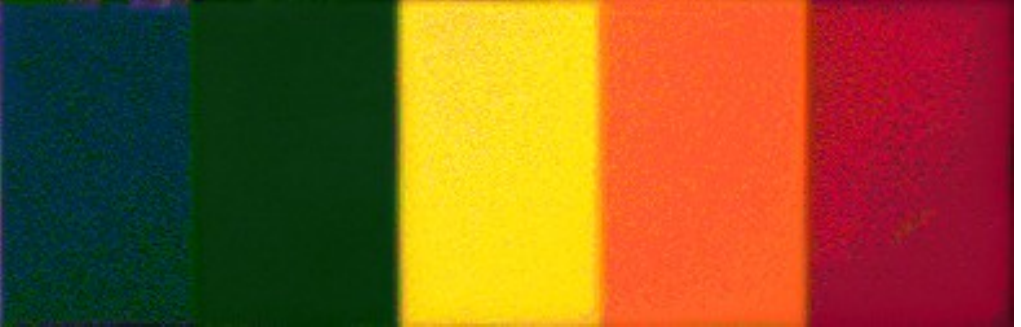}
	\caption{The icon used for real-world testing in the second set of experiments.  The true reflectance spectrum was measured for each unique section of the icon and then compared to the predicted spectra from the three inverse methods.}
	\label{fig:icons}
\end{figure}

\begin{figure}[t!]
	\centering
	\subfloat{\includegraphics[width=0.8\linewidth]{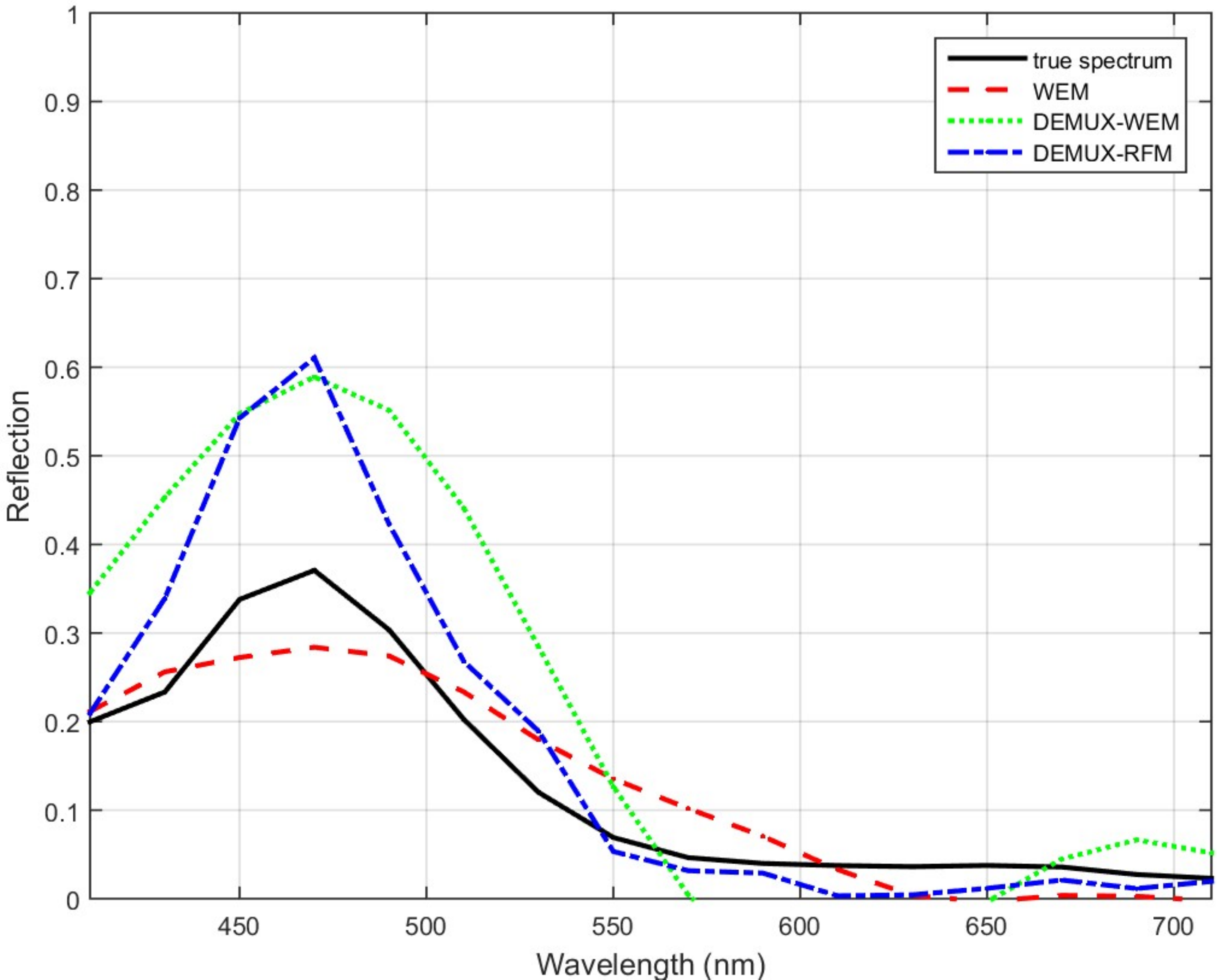}\label{01_blue}}

	\subfloat{\includegraphics[width=0.8\linewidth]{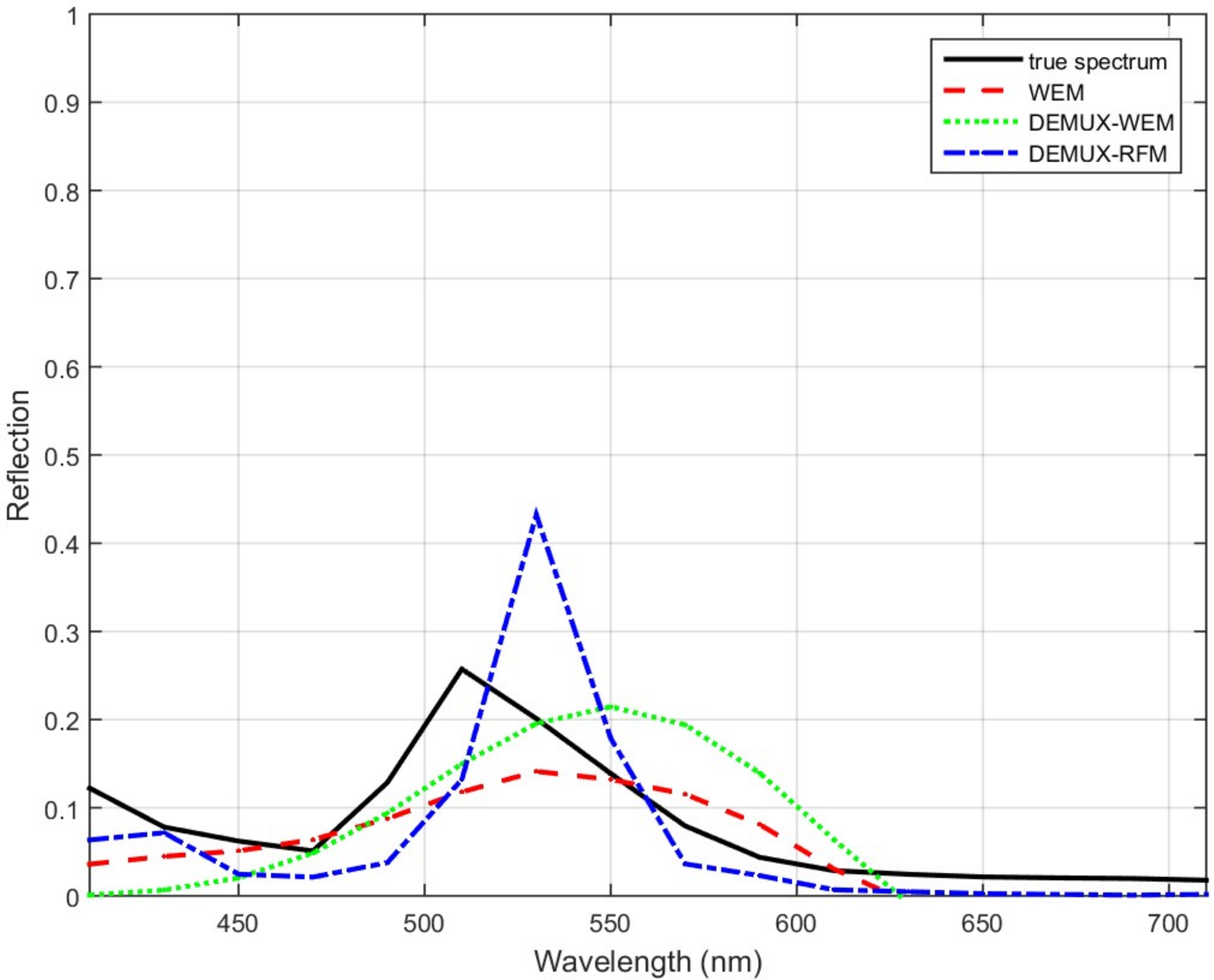}\label{02_green}}	
	
	\caption{Two of the five true reflectance spectra ('blue' and 'green') from the test icon (see Fig.~\ref{fig:icons}), along with the corresponding predicted spectra obtained from the state-of-the-art Wiener Estimation Method (WEM), the proposed Wiener-based demultiplexer (DEMUX-WEM) and the random forest-based demultiplexer (DEMUX-RFM).  Top:  The 'blue' true spectrum and the predicted spectra produced using the inverse methods have similarly shaped spectral curves. Bottom: the predicted spectra from DEMUX-RFM is closest to the true 'green' spectrum.}
	\label{fig:realspectra}
\end{figure}

\section{Experimental Setup}
\label{sec:setup}

To evaluate the efficacy of the proposed framework for performing simultaneous multispectral imaging using conventional image sensors with CFAs via numerical demultiplexing of sensor measurements, we perform two sets of experiments.  In the first set of experiments, we wish to perform comprehensive quantitative performance assessment of the proposed framework within a controlled simulation environment.  More specifically, a simulated sensor is constructed based on the characterization of the Canon T3i sensor with a Bayer pattern CFA, and a total of 10000 new randomized simulated reflectance spectra was then generated and captured using the simulated sensor to generate sensor measurements.  These measurements are then fed into the numerical demultiplexer to obtain predicted reflectance spectra.  The predicted reflectance spectra was then compared quantitatively against the original reflectance spectra entering the sensor using the peak signal-to-noise ratio (PSNR) to assess the fidelity of the demultiplexed spectra.

In the second set of experiments, we wish to assess the proposed framework within a real-world environment.  To accomplish this, we used the real spectrally-characterized Canon T3i sensor with a Bayer pattern CFA to capture sensor measurements of a test icon (see Fig.~\ref{fig:icons}).  The sensor measurements are then fed into the numerical demultiplexer to obtain predicted reflectance spectra.
The demultiplexed reflectance spectra from the numerical demultiplexer were then compared quantitatively against the known reflectance spectra of the icon using PSNR to assess the fidelity of the demutiplexed spectra.  As a baseline, the state-of-the-art Wiener estimation (WEM) method~\cite{stigell2007Wiener} was also evaluated alongside the proposed Wiener-based demultiplexer (DEMUX-WEM) and random forest-based demultiplexer (DEMUX-RFM), using the same procedures for both sets of experiments.

The true reflectance spectra of each section in the test icon was determined by measuring the sections using a high-resolution spectrometer while being illuminated by a Halogen-Tungsten (2650k) broadband light source under a 45$^\circ$ - 0$^\circ$ receiver-source setup, then detrending the measured spectra by the reflectance spectrum of the light source using a 99\% reflectance target.

\section{Experimental Results and Discussion}
\label{sec:setup}

The PSNR of WEM, DEMUX-WEM, and DEMUX-RFM for the first set of experiments were 14.7dB, 17.8dB and 20.16dB, respectively.  The proposed DEMUX-WEM achieved a significant PSNR improvement over the traditional WEM, with the proposed DEMUX-RFM exhibiting significant PSNR improvements over the other two methods.
This illustrates that the efficacy of the proposed DEMUX-WEM and DEMUX-RFM is providing more generalizable approaches for predicting a greater diversity of reflectance spectra.  Two example test reflectance spectra (with top spectra exhibiting unimodal shape and bottom spectra exhibiting bimodal shape) are shown in Fig.~\ref{fig:virtualspectra}, along with the predicted spectra from the tested methods.  In both cases, the proposed DEMUX-RFM provided the most accurate predicted spectra, followed by DEMUX-WEM and then WEM.

The PSNR of WEM, DEMUX-WEM, and DEMUX-RFM for the second set of experiments are 17.7dB, 13.3dB, and 17.2dB, respectively.
While WEM achieves the highest PSNR in this set of experiments, it is important to note that the primary reason why WEM is able to achieve this level of performance is that the true reflectance spectra of the sections in the test icon very closely resembles the spectra of color patches in the Macbeth chart, which are used to train WEM as per~\cite{stigell2007Wiener}.
Nevertheless, it is very interesting to observe that the proposed DEMUX-RFM, which is constructed based on the forward model of the characterized sensor, is able to achieve a PSNR that is very close to the WEM PSNR, with a difference of just 0.5dB, which illustrates the strength of the proposed framework.

The true spectra and predicted spectra from the test methods for the 'blue' and 'green' sections of the test icon are shown in Fig.~\ref{fig:realspectra}.  When predicting the spectrum of the 'blue' section, all three methods exhibited similar performance.  However, when predicting the spectrum of the 'green' section, WEM and DEMUX-WEM exhibited similar performance while DEMUX-RFM achieved a more accurate prediction.

The results of these two sets of experiments demonstrate the efficacy of the proposed framework for enabling simultaneous multispectral imaging using conventional image sensors with standard CFAs.  Future work involves investigating alternative models for constructing the numerical demultiplexer, the integration of a more comprehensive forward model.

\vspace{-2mm}
\section{Acknowledgment}
This work was supported by the Natural Sciences and Engineering Research Council of Canada, Canada Research Chairs Program, and the Ontario Ministry of Research and Innovation.

\vspace{-2mm}
\section{Author Contributions}
J.D., F.K., and A.W. conceived and designed the concept. J.D. and A.W. designed the numerical demultiplexer. F.K. designed the spectral characterization system. J.D. and F.K. performed the spectral characterization. J.D. and F.K. performed the experiments. J.D., F.K., D.C., and A.W performed the data analysis. All authors contributed to the writing and editing of the paper.

\end{document}